%% file: main.tex
\documentclass{article}

\PassOptionsToPackage{numbers, compress}{natbib}

\pdfoutput=1
\usepackage[preprint]{nips_2018}

\usepackage[utf8]{inputenc} 
\usepackage[T1]{fontenc}    
\usepackage{hyperref}       
\usepackage{url}            
\usepackage{booktabs}       
\usepackage{amsfonts}       
\usepackage{nicefrac}       
\usepackage{microtype}      

\usepackage{graphicx}
\usepackage{color}
\usepackage{algorithm,algpseudocode}
\usepackage{enumitem}
\usepackage{cite}
\usepackage{amsmath}
\usepackage{amssymb}

\title{Robust Tumor Localization with Pyramid Grad-CAM}

\author{
  Sungmin Lee$^{1}$, Jangho Lee$^{1}$, Jungbeom Lee$^{1}$, Chul-Kee Park$^{2}$, and Sungroh Yoon$^{1}$\thanks{To whom correspondence should be addressed.} \\
  $^{1}$Electrical and Computer Engineering, Seoul National University, Seoul 08826, Korea \\
   $^{2}$Neurosurgery, Seoul National University Hospital, Seoul 110-744, Korea \\
    \texttt{sryoon@snu.ac.kr} \\
}

\begin{document}

\maketitle

\begin{abstract}
\input{abstract.tex}

\end{abstract}
\section{Introduction}\label{s:introduction}
\input{introduction.tex}

\section{Methods}
\input{methods.tex}

\section{Experimental Results}\label{s:experiment}
\input{Experiments.tex}

\section{Discussion}
\input{Discussion.tex}

\section{Conclusion}
\input{Conclusion.tex}

%

\bibliographystyle{IEEEtranN}
\bibliography{nips_2018}

\end{document}

%% file: abstract.tex

A meningioma is a type of brain tumor that requires tumor volume size follow ups in order to reach appropriate clinical decisions.
A fully automated tool for meningioma detection is necessary for reliable and consistent tumor surveillance.
There have been various studies concerning automated lesion detection.
Studies on the application of convolutional neural network (CNN)-based methods, which have achieved a state-of-the-art level of performance in various computer vision tasks, have been carried out. However, the applicable diseases are limited, owing to a lack of strongly annotated data being present in medical image analysis.
In order to resolve the above issue we propose pyramid gradient-based class activation mapping (PG-CAM) which is a novel method for tumor localization that can be trained in weakly supervised manner.
PG-CAM uses a densely connected encoder-decoder-based feature pyramid network (DC-FPN) as a backbone structure, and extracts a multi-scale Grad-CAM that captures hierarchical features of a tumor.
We tested our model using meningioma brain magnetic resonance (MR) data collected from the collaborating hospital. In our experiments, PG-CAM outperformed Grad-CAM by delivering a 23 percent higher localization accuracy for the validation set. 

%% file: introduction.tex
A meningioma is one of the most common primary brain tumors in adults \citep{mirimanoff1985meningioma}. Thanks to its benign nature, the majority of meningiomas grow very slowly, and are diagnosed incidentally without any symptoms. Therefore a ``wait and see'' policy is recommended to asymptomatic meningioma patients, until there is solid evidence of radiological growth of the tumor or it becomes symptomatic. 
However, it is sometimes difficult to discern the growth of the tumor from sequential images taken each year without any conscious effort. 
A volumetric analysis of the tumor is thus indispensable for understanding the growth rate of the individual tumor, which is required to reach appropriate clinical decisions. 
For an accurate volume measurement, it is necessary to detect the exact position of a tumor.
However, manual or semi-automatic detection of tumor is not only inconvenient in daily clinical practice, but also error-prone. 
Thus, the development of a fully-automated tool for meningioma detection is expected to have a significant impact on the clinical community if it can guarantee fast and reliable results.

In recent years, deep learning has been extended into a variety of applications in medical image analysis, such as; diagnosing diseases, detecting the locations of lesions, and segmenting them \citep{litjens2017survey,min2017deep,lee2015fingernet}. 
However, such studies are often limited to specific diseases for which benchmark data exists owing to a lack of strong annotation data often encountered in applying deep learning to the real world \citep{wang2017chestx}.
In particular, to perform tasks such as object detection and semantic segmentation strong supervision at the human level (bounding box or pixel-wise annotation) is required \citep{hong2017weakly,lee2018hicomet}.
Because medical imaging analysis requires expertise in data annotation, unlike for regular images, it is more difficult to obtain strong annotation data.
To avoid the need for full supervision, various approaches have been proposed that require only image-level category data for training (weak annotation data). 
One such technique is localization using a class activation mapping (CAM) \citep{zhou2015learning}.

CAM performs localization based on the following principles: by applying average pooling to the entire feature map of each unit directly after the final convolution layer, which is called global average pooling (GAP) \citep{lin2013network}, we preserve the spatial information within the feature map. Afterwards we visualize the class-relevant discriminative region(s) in the input image \citep{park2017deep}. 
Target objects can be detected by drawing a tight box around the region.
This technique is effective, but also has some limitations. A CAM represents the information from the last layer of a convolutional neural network (CNN) (which should pass through many convolution layers), and has a much lower resolution than the original image.
We can only detect an approximate location of an object in the CAM, and some details might have been lost. 
To address this issue, Grad-CAM, which is a generalized version of CAM, have been proposed \citep{selvaraju2016grad}.
However, a performance improvement to capture the shape of the target object is still required.


An important observation is that the information extracted by each layer in a CNN possesses different characteristics \citep{kwon2017deepcci}.
In lower layers, we can extract low-level features such as lines and edges, when moving to the upper layers we can observe high-level patterns (such as wheels or windows in an automobile) by combining the features from the lower layers \citep{zeiler2014visualizing}.
A feature pyramid network (FPN) \citep{lin2017feature} is a model in which the CNN architecture is designed using an encoder-decoder type pyramid structure so that the feature maps of each layer are intended to form the above-mentioned hierarchical characteristics.
Feature maps extracted from each layer of an FPN can be combined with a detection or segmentation model to the further improve performance.
Therefore, we consider a multi-scale CAM approach that can exploit these layer-specific properties of CNN a plausible solution.
Different instances of the CAM extracted from different layers will be able to complement one another, often representing the discriminative details of objects to be localized.

In a similar approach, a method of extracting a multi-scale CAM using multiple GAPs has been attempted in previous work \citep{feng2017discriminative}, but during the learning process each branch (GAP with skip connection) that connects various layers and a classifier causes interference degrading the classification and localization performance.
We applied a dense connection \citep{huang2017densely} to effectively transfer multi-scale feature maps generated from an FPN to a classifier without using multiple GAPs.
A dense connection prevents all information in the lower layer from disappearing, even if the depth of the layer is increased by connecting all the layers with a direct connection.

In this paper, we propose a novel approach to tumor localization based on a pyramid gradient-based class activation map (PG-CAM). The proposed PG-CAM approach can be trained using weakly annotated data. Our model consists of densely connected layers, each of which transfers multiple scale information without loss. Consequently, PG-CAM can generates CAMs from various scales. These outputs are aggregated in the final map (i.e., PG-CAM), which can  robustly capture tiny details of an object that would otherwise be missed using an original CAM. The proposed PG-CAM model does not require time-consuming processing to generate patches by using a window-based selective search, significantly expediting the inference procedure.

Our specific contributions include the following:
\begin{itemize}
\item We propose a pyramidal fusion of gradient-based class activation maps, with the final results culminating in a single map called the \textit{pyramid gradient-based class activation map} (PG-CAM). The proposed model only needs weakly annotated data for training. In our experiments, this PG-CAM approach delivered superior performance compared to the existing single-step Grad-CAM approaches compared. 
\item We evaluate the effectiveness of using multi-scale features in a CNN and present results from various ablation experiments. In addition, we show the feasibility of feature pyramid network as a classification model. 
\item For effective training, the proposed method use the batch normalization directly after concatenation operation performed, because dense connection and concatenation operation can affect the distributions of activated value from various layers. 
\end{itemize}

\begin{figure*}[t]
  \centering
  \includegraphics[width=1.0\linewidth]{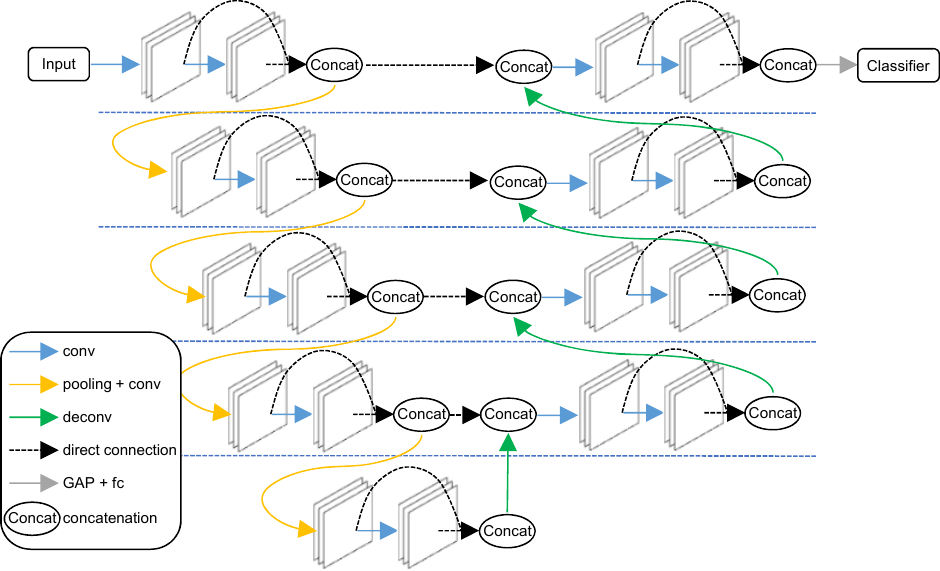}
  \caption{Overview of the proposed approach. The blue, yellow, green, and gray arrows indicate convolution, convolution after max pooling, deconvolution, and the fully-connected layer after GAP, respectively. The black dotted arrows represent the direct connection, and the dotted blue line separates the scale of the network.}
  \label{fig_overview}
\end{figure*}

%% file: methods.tex
Here, we describe in further detail the proposed P-CAM approach, which is motivated by the single CAM architecture for object localization tasks\citep{zhou2015learning} and FPN\citep{lin2017feature} for multi-scale feature extraction. 
Fig.~\ref{fig_overview} illustrates the PG-CAM architecture, which consists of an FPN component and a CAM component. 
The dense connection of the FPN's respective layers allows for the feature map, which corresponds to all scales of the feature pyramid, to directly be delivered to a classification layer.
Therefore, PG-CAM becomes an activation map in which all scale features of the feature pyramid are integrated.
%
\subsection{Architecture and Training}
\paragraph{\textbf{Grad-CAM:}}
CNNs are capable of representation learning that can extract significant features from the data. To understand this capability, there have been studies analyzing the visual encoding process of CNNs. The notion of CAM~\citep{zhou2015learning} originated from such efforts. CAM can highlight the regions identified by a CNN for performing classification.
In traditional CNNs, the fully connected (FC) layers can be used to formulate relationships between elements appearing in feature maps. However, in this process spatial information is preserved until directly after the last convolution layer can be lost.

To resolve this issue, a GAP layer is inserted directly before the fully connected layer~\citep{lin2013network}.
Then, the input to the softmax function for a category $c$ and unit $k$ can be converted to a CAM by using simple distributive law. CAM is described as follows: 
\begin{align}
CAM_{c} = \sum_{k} w_{c}^{k} f_{k}(X,Y) \label{eq:cam}
\end{align}
where $f_k(X,Y)$ represents the $k$-th feature map in the last convolution layer at the spatial location $(X,Y)$.

Grad-CAM \citep{selvaraju2016grad} is a generalized version of CAM. 
The biggest difference between Grad-CAM and CAM is that Grad-CAM uses the gradient $\frac{\partial L^c}{\partial f_k}$ of the feature map $f_k$  with respect to the class loss $L^c$ when calculating the weight for the feature map. The assumption in the Grad-CAM is that the gradient associated with the class can act as a weight. Thus, it is possible to extract the CAM not only from the feature map immediately before the GAP layer, but also it extract from the existing all feature maps of the convolution layer.
Grad-CAM is expressed by following equation EQ.\ref{eq:gradcam}. As describe in EQ.\ref{eq:gradcam}, weight values are obtained by taking an average for a gradient existing in each feature map.

\begin{align}
GradCAM_{c} = ReLU(\sum_{k} \sum_{i}\sum_{j} \frac{\partial L^c}{\partial f^k_{ij}} f_{k}(X,Y)) \label{eq:gradcam}
\end{align}
where $i$ and $j$ denote the spatial location of gradient in feature map $f_k$.
In Grad-CAM, ReLU activation is applied to the extracted CAM. ReLU activation is used to select information activated with a positive value, because of a Grad-CAM may have many negative activated values.
In this study, we propose a method of extracting CAM more precisely than previous methods by utilizing the advantages of Grad-CAM which enables CAM extraction from any layer and the advantages of encoder-decoder structures.


\paragraph{\textbf{Densely Connected Feature Pyramid Network:}}
The proposed densely connected feature pyramid network (DC-FPN) consists of a CNN-based encoder-decoder. 
This is a network designed to form a hierarchical feature map which is represented in each layer.
The encoder and decoder of DC-FPN consists of a symmetrical structure with the same number of channels and stride, and are connected by a direct connection on the same scale so that structural characteristics are preserved on each scale.
The overall structure looks similar to a U-net \citep{ronneberger2015u}, but all layers are connected by a dense connection, and the number of channels in each layer has also been modified accordingly.
As shown in Fig.~\ref{fig_overview}, at the joint where a dense connection is applied, feature maps are concatenated, and batch normalization (BN) \citep{ioffe2015batch} is applied to adjust the distribution of the activated values.
 The feature maps of the previous layer preserved by the dense connection are transferred to the final layer of the network.
Although, U-net, which is the backbone architecture of FPN, is not a suitable structure for classification, the proposed method using dense connection that exploits concatenation and BN can successfully train it.
Experimental results related to this effect will be presented in the ablation study of Table. \ref{Cls_error_mng}.


\paragraph{\textbf{Pyramid Grad-CAM:}}
The principle of obtaining a Pyramid Grad-CAM is expressed mathematically as follows: 
it is difficult to accurately track the information flow of the multi-scale features because the number of channels for feature maps is reduced during the upsampling process. For clarification, we thus represent the concept of our proposed method with a simple equation.
The symbol for the feature maps at the end of each scale is denoted as $\{f^{s_p}\}^{n}_{p=1}$. $p$ and $n$ are an index of the scale and the number of scales, respectively.
Dense connections have the effect that each feature map is concatenated, resulting in the feature maps being passed to the classifier as $concat(f^{s_1},...,f^{s_n})$.
Therefore, PG-CAM can be described as in Eq.~\ref{eq:pcam},

\begin{align}
PGCAM_{c} = \sum^{5}_{p = 1} ReLU(\sum_{k_p} \sum_{i}\sum_{j} \frac{\partial L^c}{\partial f^{k_p}_{ij}} f^{s_p}_{k_p}(X,Y)) =  \sum^{5}_{p = 1} GradCAM_{c}^{s_p} \label{eq:pcam}
\end{align}

where $k_p$ is the unit of the feature map $s_p$, $GradCAM_{c}^{s_p}$ is a GradCAM generated from the feature maps delivered from each scale $p$, and the PG-CAM aggregating the GradCAMs will consequently contain information on multiple scales of the feature pyramid.
The dense connection of PG-CAM reduces the classification performance degradation, as shown in a previous study~\citep{feng2017discriminative} that tried to extract a CAM for a multi-scale feature map by using multiple GAPs, where the performance boost achieved by adopting PG-CAM was about 8.7 percent point.
In addition, it will be shown by our experimental results that using PG-CAM in the tumor localization task is effective for capturing subtle details of the target object, which are missed for a single CAM.

As mentioned in the above paragraph, the PG-CAM merges the information of the lower scale layer during the upsampling process and transfers it to the feature map of the next scale. Therefore, it is difficult to trace exactly which features affect the classification result. In addition, even if the information in the lower layer is sent through a concatenation operation, the effect of the feature at the smallest size can be weakened. (The details about this issues will be discussed in Section \ref{s:influence}.)
On the other hand, Grad-CAM has a property to visualize the feature map associated with the class from any convolutional layers of the network using gradient generated by backpropagation.
In this paper, we also propose a complementary integration of multi-scale PG-CAM.  

\paragraph{\textbf{Loss Function and Optimization:}} To perform localization, PG-CAM only requires image-level annotations of whether a tumor exists. Therefore, we train our model with a softmax regression loss for binary tumor classification.  
PG-CAM exploits ADAM optimization~\citep{kingma2014adam}.
At this time, $\beta_{1}, \beta_{2}$ and $\epsilon$ were 0.9, 0.999 and $10^{-8}$, respectively. The base learning rate started at 0.001 and was multiplied by 0.1 for every 10,000 iterations.
 
\paragraph{\textbf{Object Localization:}}
A discriminative region in the input image tends to have a high CAM activation score between zero and one. We empirically set the threshold for detecting a tumor in the PG-CAM to 0.4 of the CAM activation score and the ResCAM to 0.8 of the CAM activation score. We represent each localized tumor using a bounding box.

%% file: Experiments.tex
We used a brain magnetic resonance (MR) image dataset of 550 patients with meningioma, collected from a hospital to which one of the authors belonged.
We performed our presented experiments using a Linux machine running CentOS, equipped with four P100 (GDDR5 15GB) GPU cards. All experiments are implemented on the Caffe \citep{jia2014caffe}.
\subsection{Meningioma MR Dataset}
This study was approved by the institutional review board of a collaborating hospital, who waived informed consent.
Each patient underwent up to 12 follow-up examinations, and the resulting images were composed of slice images of axial, coronal, and sagittal axes with T1 modality.
The training data used to train the classification model consists of 171107 image slices from 500 patients, and the validation data consists of 18322 image slices of the remaining 50 subjects. All image slices are resized and center-cropped to fit the receptive field size (224 $\times$ 224 pixel lengths).
Two neurosurgeons read all the image slices, and annotated the presence of tumors.
The ratio of the image slices with tumors to without tumors is imbalanced at approximately 85:15.
In order to measure the localization performance, regions where a tumor was located were annotated in the form of a bounding box on 188 axial images for seven patients in the validation set.
\begin{table}[t]
\caption{Classification results for the meningioma MR validation set. We show the ablation study results to demonstrate the effect of the feature pyramid network (FPN) and dense connection (DC). A comparison of the performance of PG-CAM and Multi-GAP CAM showed the effectiveness of dense connection. ResCAM (Grad-ResCAM) showed the best classification performance due to the number of layers, however the difference is negligibly small. Although the number of layers is relatively small, the difference in classification performance is only 0.8\% due to the effect of DC. On the other hand, in the localization experiments, the model using FPN recorded higher performance than ResCAM and proved its effectiveness \label{Cls_error_mng}}
  \centering
  \begin{tabular}{lcccc}
\toprule
  Model & FPN & DC & \# of Layers & Accuracy\\
  \midrule
  PG-CAM (Our method)& O & X & 22& 83.3\% \\
  PG-CAM (Our method)& O & O & 22& 95.6\% \\
  \midrule
  ResCAM~\citep{he2016deep} & X & X & 50& 96.4\% \\
  Grad-ResCAM~\citep{selvaraju2016grad} & X & X & 50& 96.4\% \\
  Multi-GAP CAM~\citep{feng2017discriminative} & O & X& 22& 79.7\% \\
  Multi-GAP CAM~\citep{feng2017discriminative} & O & O& 22& 82.6\% \\ 
 \bottomrule
  \end{tabular}
\end{table}

\subsection{Classification Results}
As shown in Table.~\ref{Cls_error_mng}, to confirm the effect of the feature pyramid and dense connection, an ablation study was conducted on the classification performance.The models compared in this experiment are PG-CAM, ResNet50-based single CAM (ResCAM)\citep{he2016deep}, ResNet50-based single Grad-CAM (Grad-ResCAM)\citep{selvaraju2016grad} and Multi-GAP CAM~\citep{feng2017discriminative}.
As a result of the dense connection, the classification accuracies of PG-CAM and Multi-GAP CAM are improved by 12.3 and 2.9 percent points, respectively.
In the structure with the dense connection, the accuracy of the CAM model with multiple GAPs was 13.0 percent points lower than that of the PG-CAM model. 
This is because when the skip connection is used to deliver feature maps located on each scale to a classifier, these interfere with each other during the learning process, thereby hindering a stable weight learning.
On the other hand, ResCAM without a feature pyramid and dense connection showed the best performance, with a 96.4 percent accuracy, because this model is 28 layers deeper than other models using an FPN. Although the number of layers is relatively small, the difference in classification performance is only 0.8\% due to the effect of dense connection.
Since ResCAM and Grad-ResCAM utilize the same ResNet structure, their classification accuracy is equal.
Interestingly, PG-CAM (without a dense connection) and Multi-GAP CAM (both the models with and without a dense connection) have been trained to classify most images as without tumors, due to being affected by the class imbalance of the training data, even though the achieved classification accuracy was over 80 percent.

\begin{table}[t]
\caption{Localization performance for the meningioma MR detection set. PG-CAM means the integrated result of Scale1 to Scale4. PG-CAM:Scale1 and PG-CAM:Scale1\&4 represent PG-CAM extracted from only Scale1, and PG-CAM extracted and integrated from Scale1 and Scale2, respectively. Although, ResCAM and Grad-ResCAM share the backbone structure outputs are extracted by using different equation (Eq. \ref{eq:cam} and Eq. \ref{eq:gradcam}) \label{Loc_err_mng}}
  \centering
  \begin{tabular}{lccccc}
\toprule
  Model & DC & Precision & Accuracy & F1 score\\
  \midrule
  PG-CAM (Our method)& O&0.45 & 0.34 &0.50\\
  PG-CAM:Scale1 (Our method)& O&0.42  & 0.33& 0.50 \\
  PG-CAM:Scale1\&4 (Our method)&O &\textbf{0.47}  &  \textbf{0.36}& \textbf{0.53}\\
  ResCAM~\citep{he2016deep} &X &0.13 & 0.12& 0.22\\
  Grad-ResCAM~\citep{selvaraju2016grad}& X& 0.26 & 0.24& 0.39\\
 \bottomrule
  \end{tabular}
\end{table}

\subsection{Localization Results}
We analyzed the localization performance of PG-CAM. To this end, we assessed the quality of localization by measuring the IOBB values between an extracted bounding box and the ground truth. If the extracted bounding box had an intersection over the detected bounding box ratio (IOBB) value of over 0.2, then we declared it a success. Although the IOBB value of 0.2 may be considered low, however there are cases where the threshold is adjusted for evaluation in the case of a medical image in which the boundary is ambiguous even when reviewed by humans \citep{huang2017temporal}.

\begin{figure*}[t]
  \centering
  \includegraphics[width=0.85\linewidth]{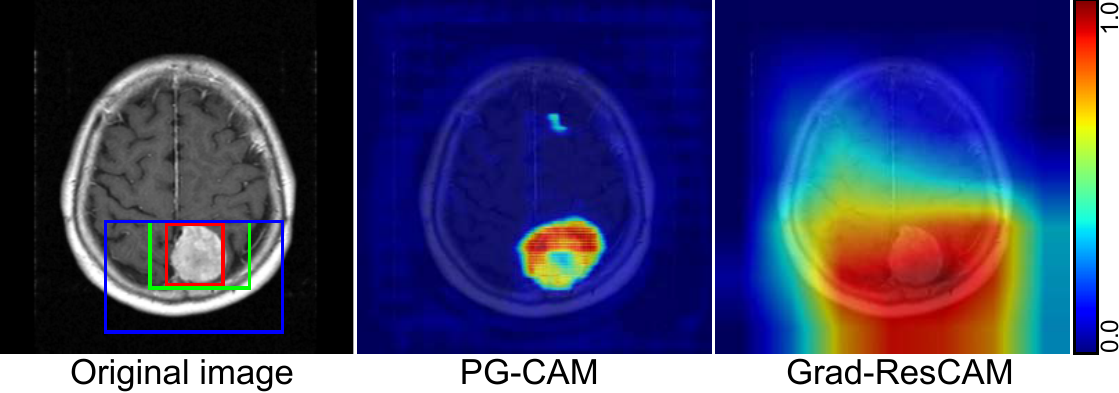}
  \caption{The qualitative results for our method. The left image shows the original input with a bounding box annotation indicating the tumor location (red). The green and blue box represent the predicted tumor locations using PG-CAM and ResCAM respectively. The other two images demonstrate the results extracted by PG-CAM (middle) and ResCAM (right). More results can be found in the supplementary material.}
  \label{fig_2}
\end{figure*}

As can be seen from Table.~\ref{Loc_err_mng}, we compared a PG-CAM model with a dense connection, which achieved the best performance among our models, with ResCAM (Grad-ResCAM), which achieved the best performance among all models.
In our settings, we tested three version of PG-CAM. PG-CAM is the fused version of PG-CAM extracted from every scales, and PG-CAM:Scale1 represents single PG-CAM extracted from Scale 1. PG-CAM:Scale1\&4 is the integration of two PG-CAM of Scale 1 and Scale 4. It is possible to combine various scales, but the reason for selectively using only the PG-CAM generated from Scale 1 and Scale 4 is because they exhibit the most different characteristics. Further details about the multi-scale characteristics of PG-CAM are covered in the Section \ref{s:influence}.
Unlike in the classification experiment, every version of PG-CAM outperformed ResCAM and Grad-ResCAM in terms of the precision, accuracy and F1 score in the localization task. 
This is because, as shown in Fig.~\ref{fig_2}, unlike ResCAM, which considers the approximate position of a tumor, PG-CAM considers multiple-scales to robustly detect the detailed position of a tumor.
In particular, the PG-CAM:Scale1\&4 showed the best localization performance, which was caused by the complementary fusion of two PG-CAMs with different characteristics.

The precision and accuracy values of the models may seem relatively lower than those of methods using strong supervision.
However, considering the fact that these two models detect the tumor without learning the bounding box annotation, and that the accuracy of the state-of-the-art weakly supervised method\citep{zhou2015learning} for the localization task for a generic image set is 43.6 percent, the localization performance of PG-CAM for the meningioma MR dataset is satisfactory.

%% file: Discussion.tex
\subsection{The Effectiveness of DC-FPN}
As can be seen from the experimental results, a very deep model with a large number of layers, such as ResNet, is advantageous for performing classification tasks. However, to visualize or detect a lesion it is advantageous to use the densely connected feature pyramid architecture, even if the depth of the network is shallow, because the details of the image to be analyzed can be robustly captured.
For example, in the case of Grad-ResCAM, the size of the CAM extracted from the last feature map is reduced to being 64 times smaller than the input image, through several pooling operations. Therefore only the approximate position is captured, as seen for Grad-ResCAM in Fig.~\ref{fig_2}.
On the other hand, the feature maps of each scale for an FPN are upsampled by the deconvolution layer to fit the size of the upper scale, and finally become the same size as the input image. Thus, the size of the extracted PG-CAM is the same as the size of the input image, and it is possible to specify the shape of the tumor in detail.
To summarize, compared with the Grad-ResCAM method, our proposed PG-CAM approach shows more robust localization performance with only a negligible sacrifice in classification accuracy.
\begin{figure*}[t]
  \centering
  \includegraphics[width=1.0\linewidth]{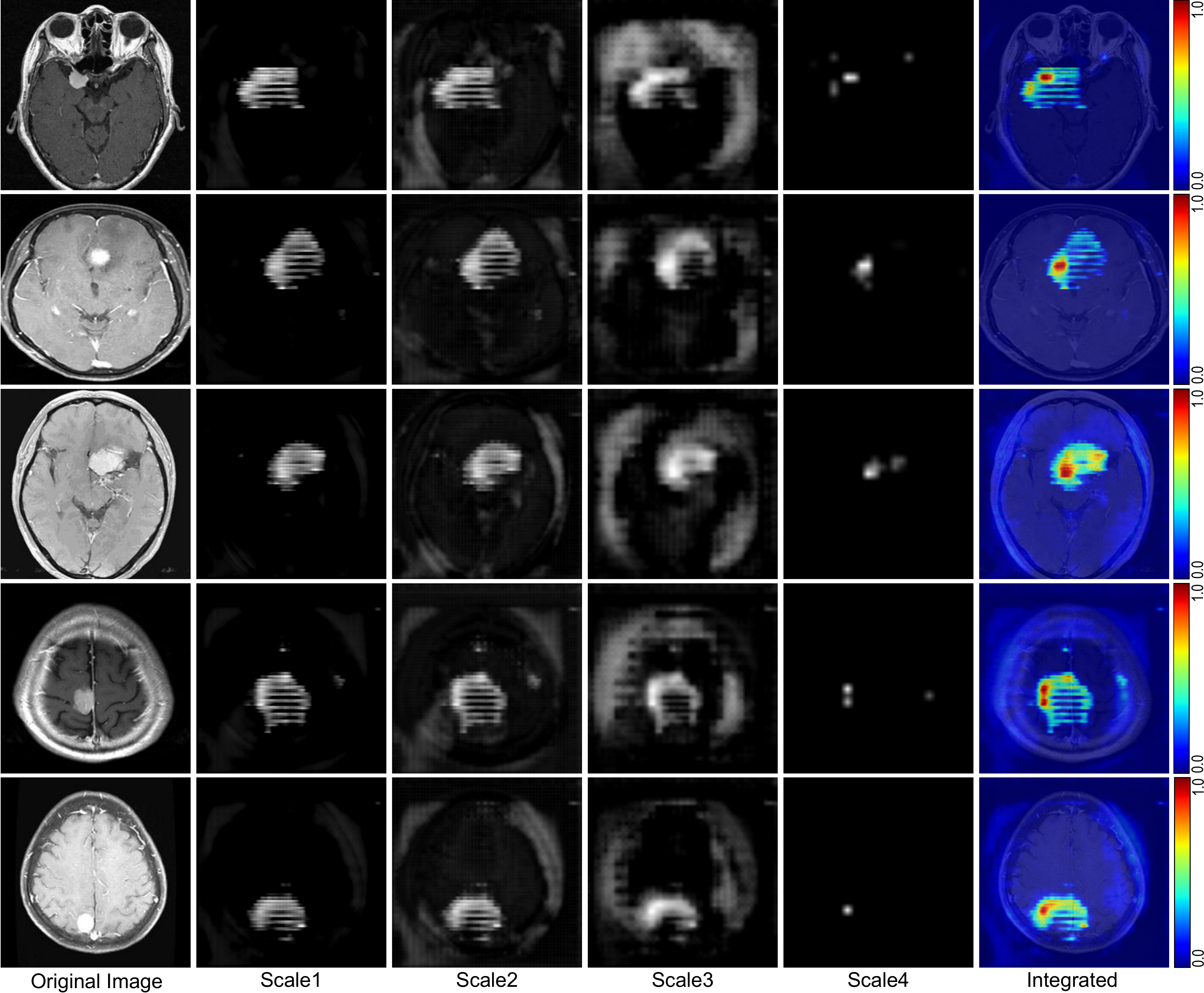}
  \caption{Examples of PG-CAM and PG-CAM extracted from each scale. Each row means the case example. Each column is sorteded in order of original input, PG-CAM extracted from four different scales and integrated version of PG-CAM.}
  \label{fig_3}
\end{figure*}

\subsection{Visualization of Multi-scale Features}\label{s:influence}
This section discusses the validity of using multi-scale features by comparing and analyzing PG-CAM extracted from different scale. 
Furthermore, we consider a combination of multi-scale features that can enhance the shape of the tumor captured by PG-CAM.
Fig. \ref{fig_3} shows the PG-CAM extracted from each scale of FPN. The outputs are illustrated in the order of the fine to coarse scale of the feature map from the left to the right. Each PG-CAM is represented by the pixel lengths of 224$\times$224, 112$\times$112, 56$\times$56 and 28$\times$28, respectively. To integrate them, we resize each of PG-CAM to the pixel lengths of 224$\times$224 (see the rightmost picture in Fig. \ref{fig_3}).
As shown in Fig. \ref{fig_3}, PG-CAM extracted from Scale 1 through Scale 3 demonstrate similar activated regions.
There is a slight difference between Scale 1 to 3: In PG-CAM Scale 1, the attention is more on the area where the tumor is located, whereas in PG-CAM Scale 2 and 3, the skull and the structural part of the brain are also activated.
Unlike the other three scales, in PG-CAM Scale 4, which captures the discriminative region on the most coarse scale, strong activation is observed in a specific part of the tumor location.
From these results, we find that the features captured on multiple scales are slightly different for the input image as described in this paper, and we can expect that the target lesion can be characterized in detail by using PG-CAM.
For this purpose, it is possible to combine PG-CAM extracted from all scales into one, and further, a method of selectively fusing scales having the most different characteristics can be considered, such as fusion of PG-CAM Scale 1 and PG-CAM Scale 4 (see localization performance of PG-CAM:Scale1\&4 in Table. \ref{Loc_err_mng}).


%% file: Conclusion.tex
In this paper, we have proposed a robust localization method named \textit{pyramid gradient-based class activation mapping} (PG-CAM). Through an ablation study, we considered and evaluated the effectiveness of the dense feature pyramid architecture from the viewpoint of tumor localization. In our experiments, the proposed PG-CAM approach delivered outstanding performances in detecting tumors in input images, achieving a 23 percent point precision boost for the meningioma MR dataset compared with the state-of-the-art single Grad-CAM method. We anticipate that the PG-CAM model, which is trainable in a weakly supervised manner, will be applicable to medical image analysis, which often suffers from a lack of annotated datasets. Furthermore, we expect that it will be possible to utilize the extracted PG-CAM as an effective and efficient preprocessing method for downstream analysis, such as in tumor segmentation.